%% file: main.tex
\documentclass{article}
\usepackage{tencent_ailab_tech_report}

\usepackage{amsmath}
\usepackage{algorithm}
\usepackage{algorithmic}

\usepackage{graphicx}
\usepackage[colorlinks=true, allcolors=blue]{hyperref}

\usepackage{subcaption}  
\usepackage[utf8]{inputenc} 
\usepackage[T1]{fontenc}    
\usepackage{hyperref}       
\usepackage{url}            
\usepackage{booktabs}       
\usepackage{amsfonts}       
\usepackage{nicefrac}       
\usepackage{microtype}      
\usepackage{soul}
\usepackage{multirow}
\usepackage{graphicx}       
\usepackage{makecell}
\usepackage{hyperref}
\usepackage{wrapfig}
\usepackage{marvosym}
\usepackage[shortlabels]{enumitem}
\definecolor{myblue}{RGB}{215,238,247}
\definecolor{mygreen}{RGB}{230,241,221}
\definecolor{mygrey}{RGB}{242,242,242}
\definecolor{myorange}{RGB}{235,142,71}
\definecolor{textgreen}{RGB}{135,201,195}
\definecolor{mypurple}{RGB}{222,213,255}
\definecolor{darkblue}{RGB}{66,141,191}
\definecolor{greyblue}{RGB}{208,220,232}

\title{Streaming Looking Ahead with Token-level Self-reward}

\author{Hongming Zhang$^1$, Ruixin Hong$^{1,2}$, Dong Yu$^1$\\
$^1$Tencent AI Lab, Seattle, $^2$Tsinghua University \\
\textit{hongmzhang@global.tencent.com, hrx20@mails.tsinghua.edu.cn, dyu@global.tencent.com}}

\begin{document}
\maketitle

\begin{abstract}

Autoregressive decoding algorithms that use only past information often cannot guarantee the best performance. 
Recently, people discovered that looking-ahead algorithms such as Monte Carlo Tree Search (MCTS) with external reward models (RMs) can significantly improve models' output by allowing them to think ahead and leverage future outputs and associated rewards to guide the current generation.
Such techniques can help the reinforcement fine-tuning phase by sampling better trajectories and the inference phase by selecting the better output.
However, their high computational cost limits their applications, especially in streaming scenarios.
To address this issue, we propose equipping the policy model with token-level self-reward modeling (TRM) capability to eliminate the need for external models and extra communication.
We name the new architecture as \textit{Reward Transformer}.
In addition, we propose a streaming-looking-ahead (\textit{SLA}) algorithm to further boost search efficiency with better parallelization.
Experiments show that \textit{SLA} achieves an overall win rate of 79.7\% against the baseline greedy decoding algorithm on three general-domain datasets with a frozen policy model while maintaining streaming efficiency.
If we combine \textit{SLA} with reinforcement fine-tuning techniques such as \textit{DPO}, \textit{SLA} achieves an overall win rate of 89.4\%. We release the experiment code at: \url{https://github.com/CognitiveKernel/SLA}.

\end{abstract}

\input{Sections/01_introduction}
\input{Sections/02_preliminaries}

\input{Sections/03_Looking_Ahead}
\input{Sections/04_method}

\input{Sections/05_experiment}
\input{Sections/07_conclusion}

\newpage
\bibliographystyle{tencent_ailab_tech_report}
\bibliography{tencent_ailab_tech_report}

\newpage

\end{document}

%% file: Sections/01_introduction.tex
\section{Introduction}\label{sec:introduction}
Language models have demonstrated remarkable capabilities in broad natural language processing tasks, from text generation to question answering~\citep{yang2019xlnet,brown2020language,ouyang2022training,chowdhery2023palm}. 
However, despite their impressive performance, autoregressive decoding algorithms that only use past information lead to suboptimal results, particularly in unseen complex tasks~\citep{song2024good}.



To further enhance LLM's performance, recent research proposes framing the generation process of large language models as a trajectory optimization problem, similar to reinforcement learning (RL), and optimizing it using advanced search algorithms like Monte Carlo Tree Search (MCTS)~\citep{metropolis1949monte,silver2016mastering}. 
MCTS allows the model to look ahead before making a move by incorporating an independent reward model (RM) to evaluate the quality of future generation trajectories.
The advanced searching techniques are crucial for both the training and the inference phases.
During the training phase, we could utilize MCTS to find an optimal solution and then upgrade the policy model through reinforcement fine-tuning (RFT)~\citep{xie2024monte,zhang2024rest,wang2024towards}.
Given a fixed policy model during the inference phase, such searching algorithms can also help the model find a better trajectory than simply using greedy decoding~\citep{zhang2022efficient,liu2024don}.


Despite its advantages, the RM-based MCTS approach has notable efficiency limitations when applied to the LLM scenario, especially in streaming ones. 
In the LLM domain, the policy and reward models typically have billions, even trillions of parameters~\citep{donisch2024inference,wang2024litesearch}, which costs not only a high computation burden but also a high communication burden since we need to distribute these models to separate machines.
Moreover, the time complexity of these search algorithms is exponential towards the trajectory length.
The overall complexity will become unacceptable if we directly apply search algorithms to the token level.
To address the challenge, existing works mostly reduce the trajectory length by using more coarse-grained action granularity, such as defining each sentence, equation, or code block as an action~\citep{xu2023no,dainese2024generating,zhang2024accessing,brandfonbrener2024vermcts}.
However, these heuristic approaches are limited to specific structured domains and do not apply to general-domain tasks.

This paper aims to unleash the power of looking ahead algorithms by removing the external reward model.
To achieve this goal, we propose to enhance the transformer architecture~\citep{waswani2017attention} with an additional channel to model rewards simultaneously.
This design offers three advantages.  
First, since the architecture can produce reward estimation with near-zero additional computational cost for each token during exploration, it supports simultaneous token-level reward modeling (TRM), greatly expanding the method's applicability to general-domain tasks.
Second, integrating generation and self-evaluation into a single model significantly reduces the communication overhead of multi-model interactions.
Third, distributing the reward modeling into each transformer block could significantly improve the reward modeling performance compared with widely used adapter-based methods.
In the rest of the paper, we denote the proposed architecture as \textit{Reward Transformer}.


\begin{figure}
    \centering
    \includegraphics[width=0.7\linewidth]{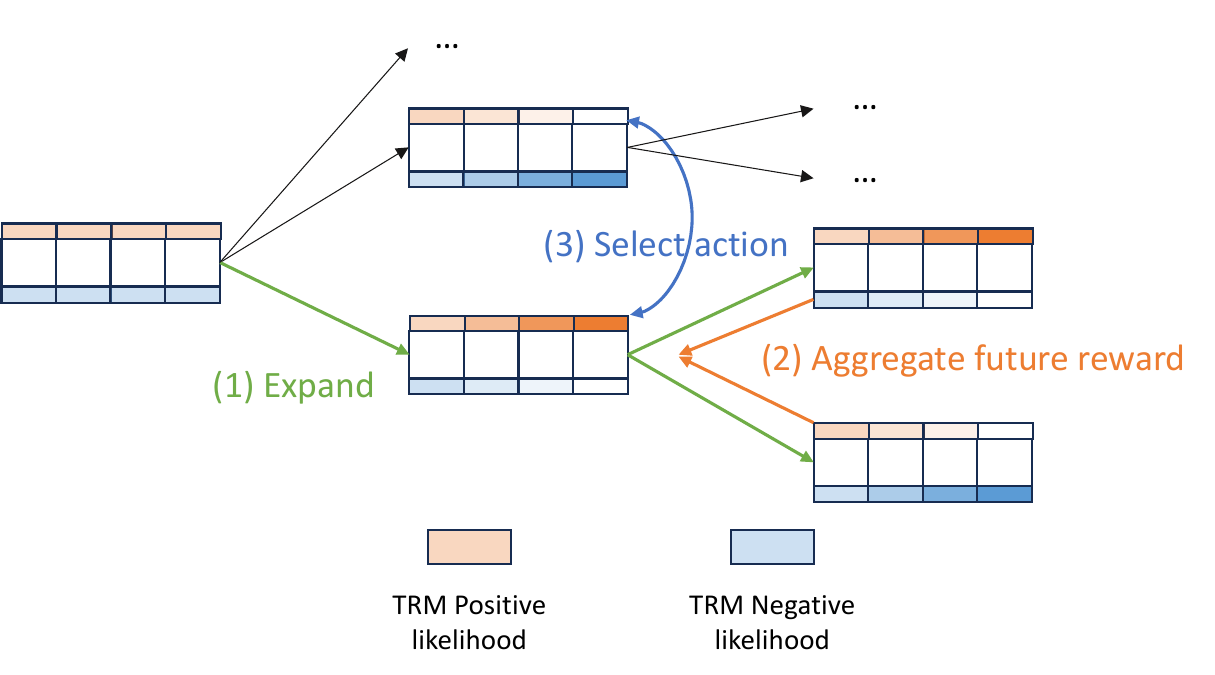}
    \caption{Overall framework of the proposed streaming looking ahead algorithm. We leverage the proposed \textit{Reward Transformer} that combines policy modeling and token-level self-reward modeling to conduct the efficient looking-ahead search. At each step, the policy model will generate future tokens and the associated reward simultaneously and then aggregate back to select the better action.}
    \label{fig:framework}
    \vspace{-0.1in}
\end{figure}

Building upon the trained model, we implement a streaming look-ahead (\textit{SLA}) algorithm, with a demonstration in~\autoref{fig:framework}. 
At each step, the policy model looks ahead by simultaneously generating future tokens and the associated reward, which it then aggregates to select the better action.
The whole process is streaming since generated future tokens can be directly reused for the next step of decision-making.
Unlike previous MCTS works that require pre-defined steps~\citep{xu2023no,dainese2024generating,zhang2024accessing,brandfonbrener2024vermcts}, \textit{SLA} supports arbitrary action granularity down to a single token to guarantee generalization capability.

Evaluation on three general-domain datasets~\citep{DBLP:conf/nips/ZhengC00WZL0LXZ23,DBLP:journals/corr/abs-2406-11939,DBLP:journals/corr/abs-2404-04475} demonstrates that \textit{SLA} achieves a 79.7\% win rate against the baseline greedy decoding with a frozen policy model.
If the RFT is allowed, the combination of \textit{SLA} and \textit{DPO} could achieve an 89.4\% win rate.
To better monitor the TRM performance, we also propose a new evaluation metric AuTRC, with more details in \autoref{sec:analysis_TRM}.
Empirical experiments show that the proposed architecture could significantly improve the TRM quality compared with existing methods.

The paper is organized as follows. In \autoref{sec:problem_formulation}, we provide the problem formulation and essential background. 
In \autoref{sec:MCTS}, we introduce the importance and limitation of existing looking ahead algorithms in LLM applications.
In \autoref{sec:method}, we introduce the details of \textit{SLA} and \textit{Reward Transformer}. In \autoref{sec:experiments}, we present empirical experiments to demonstrate the effectiveness of SLA and associated design choices.
Lastly, \autoref{sec:conclusion} concludes this paper.

%% file: Sections/02_preliminaries.tex
\section{Preliminaries}\label{sec:problem_formulation}


In this section, we first introduce how the LM generation process can be formulated as a token-level Markov Decision Process (MDP) and then explain how existing sampling algorithms relate to it.

\subsection{LLM Decoding as Token-level MDP}

The language model generation process takes a sequence of tokens as inputs and generates a sequence of tokens as outputs.
Mainstream transformer-based language models generate the output tokens one by one until the stopping criteria (e.g., an end-of-sequence token or maximum sequence length) are met.
The traditional MDP is usually formulated as a tuple $\mathcal{M} = (\mathcal{S}, \mathcal{A}, F, R, \gamma)$, where $\mathcal{S}$ is the set of all possible states, $\mathcal{A}$ is the set of actions, $F$ is the transition function, $R$ is the reward function, and $\gamma$ is the decay parameter.
In the language model scenarios, each state in $\mathcal{S}$ is a trajectory that can be denoted as $\tau$.
Each action in $\mathcal{A}$ is selecting a token $x$ from the vocabulary set.
$F$ is the deterministic transition of concatenating the selected action (i.e., a token) with the existing state (i.e., a trajectory) to become a new one.
Traditionally, rewards $r_t$ are defined at every step $t$ and contribute to the return $G_t=\sum_{k=0}^{\infty} \gamma^k R_{t+k}$ through the decay factor $\gamma$.
However, in the LLM scenario, we are only concerned with the quality of the complete trajectory generated, meaning that the reward function $R(s)$ evaluates the final trajectory rather than providing step-by-step feedback.
Thus, the return becomes $G=R_T$, where $R_T$ is the reward associated with the final sequence at step $T$, and $\gamma$ is irrelevant because intermediate rewards are not accumulated. 

\subsection{The Classical Decoding Algorithms}

Formally, given an input \( \mathbf{x}  = (x_1, x_2, \ldots, x_T) \), a reward function $R$ that provides a scalar reward for a trajectory $\tau$, and a language model \( p_\theta(x) \) parameterized by \( \theta \), the goal of decoding algorithms is to find the optimal trajectory $x^\star$ sampled from \( p_\theta(x) \) that could maximize the reward:

\[
\tau^* = \arg\max_{\tau \sim p_\theta(\mathbf{x} )} R(\tau).
\]

This section covers representative decoding algorithms and explains how they are connected.

\textbf{Greedy Decoding}: The naive but most widely used algorithm is \textit{Greedy Decoding}, which uses the language modeling likelihood at each step as guidance.
At each step $i$, this algorithm selects the action token $x_i \in \mathcal{A}$ following:
\begin{equation}
x_i = \arg\max_{x} P(x \mid x_{<i}).
\end{equation}
From the angle of MDP, this method uses the accumulative likelihood predicted by the language as the final reward:
\begin{equation}
 R(\tau) \gets \Pi_{i}^{T}P(x_i \mid x_{<i}),
\end{equation}
where $T$ is the length of $\tau$, and takes a greedy solution to approach this goal.

\textbf{Sampling-based Decoding:} 
On top of greedy decoding, people also try to incorporate diversity in the final output.
For example, the temperature-based method introduces an additional parameter $\lambda$ to control the greedy sampling process by reshaping the likelihood distribution as:
\begin{equation}
x_i \sim P(x \mid x_{<i})^{1/\lambda}.
\end{equation}
From the angle of token-level MDP, we can reinterpret this process as introducing an additional diversity objective:

\begin{equation}
 R(\tau) \gets \Pi_{i}^{T}P(x_i \mid x_{<i}) \cdot D(x_i, x_{<i}),
\end{equation}
where 
\begin{equation}
    D(x_i , x_{<i}) = P(x_i \mid x_{<i})^{\frac{1}{\lambda}-1}.
\end{equation}

To avoid sampling rare tokens and achieve a balance between performance and diversity, researchers have investigated how to dynamically adjust the candidate token pool~\citep{holtzman2019curious,zarriess2021decoding}. 
For example, the \textit{Top-$k$ Sampling} algorithm only considers the top $k$ tokens with the highest probabilities as candidates instead of the whole vocabulary.
Similarly, the \textit{Nucleus Sampling}, which is also known as \textit{Top-p Sampling}, only selects from the smallest possible set $\mathcal{V}_p \subseteq \mathcal{V}$, where the cumulative probability mass exceeds a threshold $p$.


\textbf{Trajectory-level Decoding:} Although these token-level decoding algorithms are efficient, they tend to generate locally coherent outputs that may lack global quality.
To solve this problem, people also developed decoding algorithms that consider partial or whole trajectories.
For example, the \textit{Beam Search Decoding} algorithm keeps track of the top $B$ partial trajectories, expanding them at each step and retaining only the ones with the highest joint likelihood.
Similar to the \textit{Greedy Decoding}, this method also uses the joint likelihood as the trajectory reward function.

\textbf{Advanced Reward Modeling Algorithms:}
A common limitation of the aforementioned algorithms is their fundamental assumption that the joint likelihood could represent $R(\tau)$ might not always hold.
People have been interested in introducing better reward signals as guidance to address this.
For example, in the QA scenario, the \textit{Majority-voting algorithm} assumes that the more frequent answer aligns better with the grounding reward function (i.e., accuracy) and thus selects candidate trajectories following this guidance.
Though this intuitive approach has been shown to be effective on tasks such as QA and math problems, it is restricted to tasks with structured output for voting. It cannot be generalized to more general-purpose applications.
To address this issue, researchers also include an external model $R^\prime$, which is often another transformer-based model, to approximate the ground truth reward model $R$. 
With that, we could sample $K$ trajectories $\mathcal{T}_K$ with sampling-based decoding algorithms and then use $R$ to select the trajectory with the maximum reward:
\begin{equation}
    \tau^\star = \arg \max_{\tau \in \mathcal{T}} R^\prime(\tau).
\end{equation}
Employing an external model to model the reward offers greater flexibility than heuristic rewards. This approach is not constrained by the structured answer format, which improves generality and adaptability in various scenarios.

%% file: Sections/03_Looking_Ahead.tex
\section{Looking Ahead Search Algorithms}
\label{sec:MCTS}

A critical limitation of the aforementioned approaches is that they rely solely on past information, which may not be sufficiently informative for making a wise decision. 
To address this, researchers try to enable the model to look ahead and revisit its choices to make a more informed decision. 
This methodology mirrors how humans plan ahead before making a decision.
Such ideas were widely used in the traditional RL tasks such as the GO game~\citep{silver2016mastering,silver2017mastering} and one of the most widely used algorithms is the Monte Carlo Tree Search (MCTS)~\citep{metropolis1949monte}.

As shown in Algorithm \autoref{algorithm:MCTS}, when making a move, a typical MCTS algorithm typically involves the following steps: (1) selection: following the UCB policy to find a leaf node; (2) expansion: expand it if the located node is not the final state; (3) simulation: calculate the reward for the current state; (4) backpropagation: update the Q-value and visit count for all previous nodes.
This expand-simulation-backpropagation procedure is essentially a way of looking ahead and using future information for the current decision.
Typically, we repeat this procedure for $N$ times/rollouts and then decide based on the future rewards collected.

Recently, MCTS has been introduced into the LLM scenario for both the training and inference stages~\citep{zhang2022efficient,xie2024monte,zhang2024rest,wang2024towards,liu2024don}.
During training, MCTS was known as a power algorithm for sampling good responses, which can be used to optimize the model.
On the other hand, search methods like MCTS have also been proven to be a powerful inference algorithm for improving the model's performance on complex tasks~\citep{feng2023alphazero,lightman2023let}.
However, as the Deepseek technical report~\citep{deepseekai2025deepseekr1incentivizingreasoningcapability} discussed, efficiency is still the Achilles' heel of applying MCTS in LLM.

\begin{algorithm}[t]
\small
\caption{Monte-Carlo Tree Search (Single Step)}
\label{algorithm:MCTS}
\textbf{INPUT}: Policy Model $P$, Reward Model $R$, Root Node $s_0$, Max Iterations $N$ \\
\textbf{OUTPUT}: Optimal Action $a^*$ \\
\begin{algorithmic}[1]
    \STATE \textbf{INITIALIZE}: Create a search tree with root node $s_0$ and initialize $Q(s, a) \leftarrow 0$, $N(s, a) \leftarrow 0$ for all states and actions.
    \FOR{$i = 1$ TO $N$}
        \STATE \textbf{SELECTION}: Start at $s_0$, traverse the tree by choosing child nodes using the UCB policy until a leaf node $s_L$ is reached.

        \STATE \textbf{EXPANSION}: if $s_L$ is not a terminal state, add a child node $s_{L+1}$ for each possible action $a$ and estimate the prior probability $P(s_{L+1}, a)$.

        \STATE \textbf{SIMULATION (ROLLOUT)}: Calculate the cumulative reward $r$ over the trajectory: $r = \sum_{t=0}^{T} R(s_t, a_t)$.
        
        \STATE \textbf{BACKPROPAGATION}: For each node $(s_t, a_t)$ along the path from $s_L$ to $s_0$, Update $Q$-value and visit count.
    \ENDFOR

    \STATE \textbf{RETURN}: $a^* = \arg\max_{a} Q(s_0, a)$
\end{algorithmic}

\end{algorithm}

If we use $N$ and $n$ to indicate the number of sampled trajectories and number of tokens per action.\footnote{We use ``action'' because different algorithms might use different granularities such as tokens and sentences.}
To select each action, the algorithm first expands trajectories, collects rewards for each sampled trajectory, backpropagates, and selects the action.
Since the main time costs in the large language model scenario are related to language model computing and communication, we ignore another time cost for simplicity.
For each action, the computation time complexity is
\begin{equation}
   O( N \cdot (n \cdot t_{d} + 2 \cdot t_{c} + t_{r})),
\end{equation}
where $t_d$, $t_c$, and $t_r$ are the time cost for the policy model to decode a token, communication between two models, and the reward model to generate a numerical score.
And then, if we use $t_p$ to indicate the prefilling time and $T_{max}$ as the maximum trajectory length, the total time cost will become:
\begin{equation}
   O( t_p + \frac{T_{max}}{n} \cdot N \cdot (n \cdot t_{d} + 2 \cdot t_{c} + t_{r})).
\end{equation}
Given that people often use another LM with the same or larger size as the reward model, $t_r$ is often large.
In actual applications, people reduce this complexity by choosing a relatively larger $n$ and defining each action at coarser granularity, such as a sentence.
This trick makes the MCTS search slightly more affordable but also restricts the generalization capability.

%% file: Sections/04_method.tex
\section{Streaming Looking Ahead}\label{sec:method}



We present the proposed streaming looking ahead (\textit{SLA}) algorithm in Algorithm \autoref{alg:streaming_looking_ahead}.
Compared with the naive MCTS, \textit{SLA} has two upgrades.
First, we remove the requirements of pre-defined heuristic steps to increase the generalization capability.
Instead, we directly use tokens as the minimum action granularity.\footnote{Users may choose any token as the step size; for a more in-depth analysis of the impact of the step granularity, please refer to \autoref{sec:analysis_TRM}.}
Second, we remove the separate reward model by proposing a \textit{Reward Transformer} module, an advanced version of the original transformer with an additional reward channel to simultaneously judge the current trajectory's quality.
The reward channel predicts the final reward given incomplete trajectories at the token level.
Thus, we can name this process as token-level reward modeling (TRM).\footnote{A token-level reward is an estimate of the final reward rather than an absolute ground truth. We use the term ``reward'' for clarity and to align with previous work.}
This section introduces details of the proposed \textit{SLA} algorithm and the reward transformer.



\begin{algorithm}[t]
\small
\caption{Streaming Looking Ahead (Single Step)}\label{alg:streaming_looking_ahead}
\textbf{INPUT}: Joint Policy-Reward Model $J$, Current Node $s_0$, Search Depth $d$, Search Width $k$, Step size $n$. \\
\textbf{OUTPUT}: Optimal Action $a^*$ \\
\begin{algorithmic}[1]
    \STATE \textbf{INITIALIZE}: Create a search tree with root node $s_0$ and initialize $Q(s, a) \leftarrow 0$ for all states and actions.
    \FOR{$i = 1$ TO $d$}
        \STATE \textbf{EXPANSION}: Expand all leaf nodes to $k$ children's. For each child, generate the next maximum $n$ tokens unless meeting a stopping criteria.

        \STATE \textbf{SELF-EVALUATION}: Record the Reward Transformer output $R^\prime$ at the last token of each child.
        
        \STATE \textbf{BACKPROPAGATION}: For each node $(s_i, a_i)$ along the path from $s_i$ to $s_0$, Update $Q$-value as $Q(s_i, a_i) = \max Q(s_i \mid a_i, *)$, where $\mid$ is the transaction rule of the language model concatenation and $*$ indicates all children actions.
    \ENDFOR

    \STATE \textbf{RETURN}: $a^* = \arg\max_{a} Q(s_0, a)$
\end{algorithmic}

\end{algorithm}

\subsection{Streaming Looking Ahead}
Following the notations in \autoref{sec:MCTS}, we use $t_{p}$, $t_{d}$, $t_{r}$, and $t_{c}$ to indicate the time cost for prefilling, decoding a token, generating rewards for a trajectory, and communication between different models, respectively.
Compared with the naive MCTS, \textit{SLA} makes two changes.
First, \textit{SLA} removes the communication and reward modeling cost by including the self-reward functionality in the policy model.
Second, \textit{SLA} changes the random exploration algorithm with a balanced tree search.

Here, we denote the search depth, search width, and step size as $d$, $k$, and $n$, respectively.
For each step, since there is no external remodel, all the computation come from computing the future tokens in advance. Thus, we could formulate the computation complexity as
\begin{equation}
    O(k^{d} \cdot (n \cdot t_{d})).
\end{equation}
Since the effective searched trajectory of our algorithm is $k^d$. 
Following the setting used by MCTS, we could let $k^{d} = N$ and thus rewrite $d$ as $\log_k(N)$.
Multiplying the time per step with the number of steps $\frac{T_{max}}{n}$ and the prefilling time, we can get the final time complexity as
\begin{equation}
    O(t_{p} + \frac{T_{max}}{n} \cdot (k^{\log_k(N)} \cdot (n \cdot t_{d}))).
\end{equation}

Compared with the exploration in MCTS, the main advantage of tree search is that all children from the same expansion node share the same ancestor, and we could utilize batch computing to compute all children in parallel to significantly reduce the time complexity from $k^{\log_k(N)}$ to $\log_k(N)$.
As a result, the total time complexity becomes
\begin{align}
    &O(t_{p} + \frac{T_{max}}{n} \cdot \log_k(N) \cdot n\cdot t_{d})\\
    = &O(t_{p} + T_{max} \cdot \log_k(N) \cdot t_{d}).
\end{align}
With the KV-cache technique, $t_d$ is often a small value.
Unlike the MCTS method, \textit{SLA}'s theoretical speed is irrelevant to the step size.\footnote{Pratically, frequent expansion operations could increase the time cost from other operations, such as sequence copying, in current inference engines. Hence, in our experiments, the default step size is set to 10, which is also much smaller than the sentence level used by prior works.}
As a result, \textit{SLA} can operate at a much finer granularity, down to the token level (i.e., $n=1$), making it applicable to a broader range of domains.
Compared with the greedy decoding, the prefilling time remains the same, and the decoding time changes from $t_d$ to $\log_k(N) \cdot t_{d}$.
Assuming we set $N$ and $k$ to be 64 and 4, which is a typical MCTS setup, \textit{SLA} will delay the streaming speed by $\log_4(64) = 3$ times.
Given that the typical speed for streaming applications humans could tolerate is five tokens/second~\citep{liu2024andes}, and the current inference speed is above hundreds of tokens per second~\citep{kwon2023efficient}, \textit{SLA} makes the complex looking ahead search possible for streaming applications.

\subsection{Architecture}

\begin{figure}
    \centering
    \includegraphics[width=0.4\linewidth]{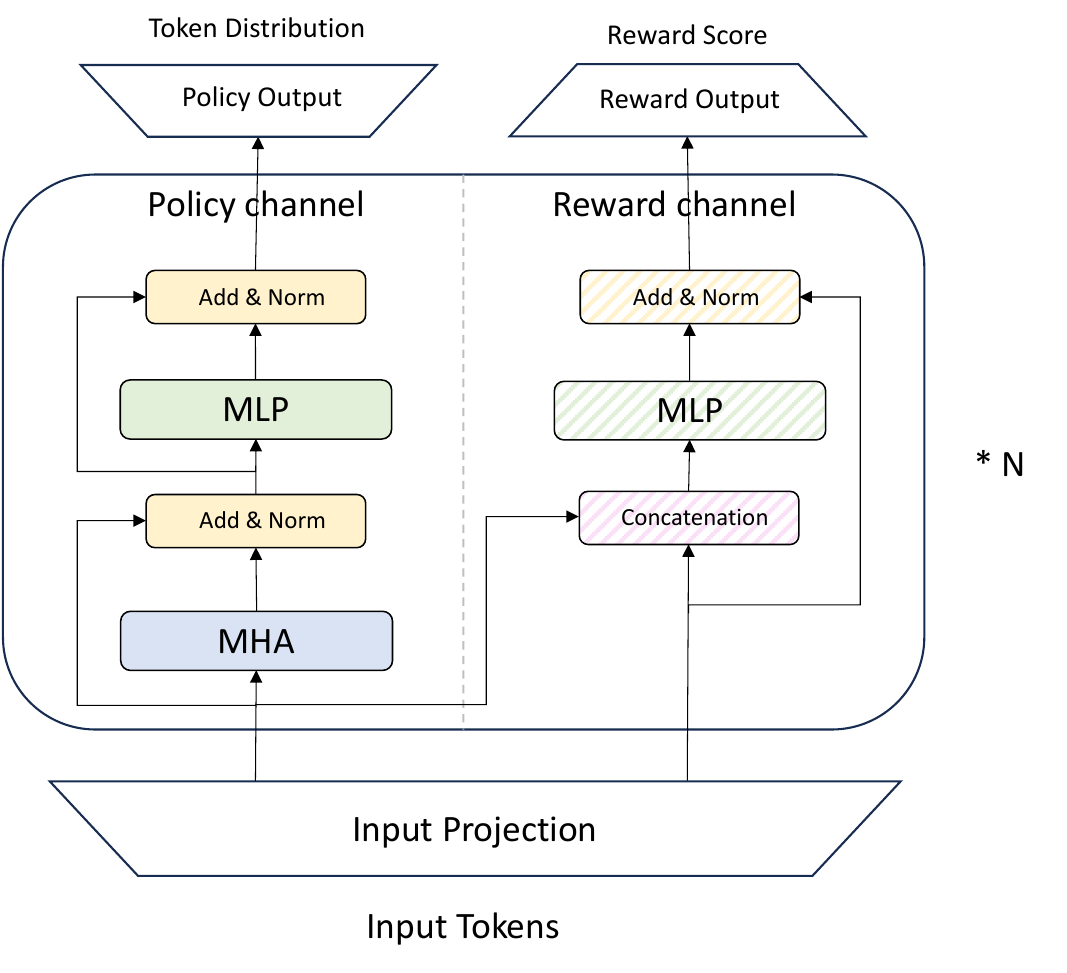}
    \vspace{-0.1in}
    \caption{Overall framework of the \textit{Reward Transformer}. }
    \vspace{-0.1in}
    \label{fig:architecture}
\end{figure}
The foundation of \textit{SLA}'s success lies in the assumption that the policy model can also function as the TRM with nearly zero additional cost.
To achieve this goal, we propose the \textit{reward transformer}.
As shown in \autoref{fig:architecture}, the network takes a sequence of tokens as input and simultaneously outputs the distribution for the next action and reward for the current input sequence.
At the bottom, we have the input projection layer to project tokens into a sequence of embedding representations.
Then, we have $N$ repeat kernels that follow a dual-channel design.
The policy channel on the left side is responsible for computing current state representation and predicting the distribution of the next actions.
The reward channel on the right side predicts the final reward given the current representations.
In the end, we have two separate output heads to project the hidden representations from two channels to distributions of the next token and rewards for the current input.

Formally speaking, we denote the input representation after the input projection as $\mathbf{X} \in \mathbb{R}^{t \times d}$, where $t$ is the number of input tokens, and $d$ is the initial token representation dimension, we first create a copy of it and then use them as the input for the first \textit{reward transformer} module.
For each layer $j$, we denote the policy and reward representation as $h_p^{j}$ and $h_r^{j}$.
We compute the policy representation for the next layer $h_p^{j+1}$ following the original transformer as
\begin{equation}
h_p^{j+1} = Norm(h_p^j+FFN(Norm(h_p^j+MHA(q_p^j, k_p^j, v_p^j)))),
\end{equation}
where $Norm$ is the normalization layer, FFN is a feed-forward layer, and MHA is a multi-head attention.
$q_p^j$, $k_p^j$, and $v_p^j$ represent query, key, and value projected from $h_p^j$.
On the other side, we compute the reward representation as:
\begin{equation}
h_r^{j+1} = Norm(h_r^j + FFN([h_p^j:h_r^j]),
\end{equation}
where $[:]$ indicates the concatanention.

We have two separate output layers after the $N$ repeated reward kernels.
For the policy channel, we follow the standard setting to get the output distribution.
\begin{equation}
    \pi(a \mid \mathbf{X}) = Softmax(W_\pi h_p^{N-1} + b_\pi),
\end{equation}
where $W_\pi$ and $b_\pi$ are output weights and biases that project the hidden representation to the whole vocabulary.
Similarly, for the reward channel, we get the reward prediction with
\begin{equation}
    R^\prime(\mathbf{X}) = W_R h_p^{N-1} + b_R.
\end{equation}

Such a design has three advantages. 
First, the policy channel inherits the original transformer, so we could utilize any policy model and only update the reward channel.
Second, compared with the adapter design, this fused design preserves the deep communications between two channels at each layer such that the reward channel could better utilize the strong language representation capability of the adopted policy channel.
Third, this fused design can be easily scaled up with bigger models.

\subsection{Optimization}

We face two challenges in optimizing the \textit{reward transformer} model.
The first is the distant supervision challenge because the annotation only applies to the complete trajectories rather than the incomplete ones. 
The second one is that people are poor at providing an absolute score for a trajectory, especially for general domain tasks such as dialogue.
Instead, they are better at comparing different trajectories.
Given these challenges and the goal of expecting the TRM functionality to work,
we propose optimizing the reward channel with the Bradley-Terry loss, where the sequence score aggregates token-level rewards.

Specifically, We formulate the training data in pairs $(\tau^w, \tau^l) \in \mathcal{D}$ to represent the winning and losing trajectories.
Tokens in the winning and losing trajectories are denoted with $a^w_i$ and $a^l_i$, respectively.
$i$ indicates the position and we use $\tau^w_i$ and $\tau^l_i$ to denote the incomplete trajectories at length $i$.
We then can define the training loss as:
\begin{equation}
    \mathcal{L} = - \mathbb{E}_{(\tau^w, \tau^l) \in \mathcal{D}} \left[ \log \sigma( \frac{1}{T_{w}}\sum_{t=1}^{T_{w}} R^\prime(\tau^w_i) - \frac{1}{T_{l}} \sum_{t=1}^{T_{l}} R^\prime(\tau^l_i) )  \right],
\end{equation}
where $T_{w}$ and $T_{l}$ is the maximum length of $\tau^w$ and $\tau^l$, respectively. $R^\prime$ is the reward output. 
We can then optimize the reward channel with the gradient descent algorithm.


%% file: Sections/05_experiment.tex
\section{Experiments}\label{sec:experiments}

We experiment with the open-sourced model {Llama-3-8B-Instruct}~\citep{DBLP:journals/corr/abs-2407-21783} and its variaces.
We set the hidden layer dimension to 256 for the reward channel.
We set $d$, $k$, and $n$ as 2, 2, and 10, respectively.
We use the ArmoRM-Llama3-8B-v0.1\footnote{\url{https://huggingface.co/RLHFlow/ArmoRM-Llama3-8B-v0.1}}~\citep{DBLP:conf/emnlp/00030X0024} to simulate the ground truth of human preference $R$.



\textbf{Training:}
We use prompts from ultrafeedback~\citep{DBLP:conf/icml/CuiY0YH0NXXL0024} dataset as our training data to train the self-reward capability of the models. 
We follow the data collection process of~\citep{DBLP:journals/corr/abs-2405-14734}.
Specifically, we first collect responses generated by the policy model and then label the responses using the ground truth reward model. 
For each input prompt, we generate five responses with a temperature of 0.8, selecting the highest-scoring response as the chosen response and the lowest-scoring response as the rejected response.
We tried two settings: (1) freeze the policy model and (2) update the policy model with DPO.
For each setting, we train for three epochs with a learning rate of 5e-5. We use the Adam optimizer and a cosine learning rate schedule.

\textbf{Evaluation:}
We follow the standard approach to use prompts from three general-domain instruction-following datasets for evaluation: MT-Bench~\citep{DBLP:conf/nips/ZhengC00WZL0LXZ23}, Arena-Hard~\citep{DBLP:journals/corr/abs-2406-11939}, and AlpacaEval 2~\citep{DBLP:journals/corr/abs-2404-04475}. 
MT-Bench encompasses 80 questions across nine categories, requiring the model to engage in two rounds of dialogue. 
Arena-Hard is an enhanced version of MT-Bench, comprising 500 well-defined queries spanning multiple domains.
AlpacaEval-2 is made up of 805 questions sourced from five datasets.
We simulate the actual application scenario to generate outputs in a streaming paradigm. As the baseline, we consider all decoding methods that are efficient enough to support streaming usage and applicable to the general domain.
Specifically, we consider naive \textit{temperature sampling} (with a temperature of 0.8), \textit{top-p} sampling (with a probability threshold of 0.9), \textit{top-k} sampling (with a threshold of 50), and \textit{beam-search} (with a beam size of 4).
Besides that, as discussed by \citep{DBLP:journals/corr/abs-2404-12358}, \textit{DPO}~\citep{rafailov2024direct} could inherently learn the human preference into the model weights such that the greedy output could perform following the reward feedback.
Thus, we also report the performance of all search algorithms combined with the model after \textit{DPO} fine-tuning.
In our experiments, we use the same paired data used by \textit{SLA} to conduct the \textit{DPO} training.
We use the same reward model to score the final outputs and compare outputs with the greedy output to calculate the win rate, which is defined as the percentage of wins plus half of the ties.

\begin{table*}[t]
\scriptsize
\centering
\caption{Performance Comparison of different approaches against the baseline greedy decoding. In the bottom line, we also directly compare ``llama-DPO + SLA'' against ``llama-DPO + Greedy'' to better understand their performance. }
\label{table:main_results}
\begin{tabular}{l|ccc|ccc|ccc|c}
\toprule
Model & \multicolumn{3}{c|}{MT\_Bench} & \multicolumn{3}{c|}{ArenaHard} & \multicolumn{3}{c|}{AlpacaEval} & Avg. win\_rate \\
 & win & tie & loss & win & tie & loss & win & tie & loss & \\
\midrule
llama-instruct + temperature &51.2 & 0.6 & 48.1 & 48.6 & 0.0 & 51.4 & 48.4 & 0.5 & 51.1 & 49.6\\
llama-instruct + top-P& 50.6 & 0.0 & 49.4 & 50.8 & 0.0 & 49.2 & 51.1 & 0.7 & 48.2 & 51.0 \\
llama-instruct + top-K& 45.0 & 0.0 & 55.0 & 50.8 & 0.0 & 49.2 & 47.8 & 0.4 & 51.8 & 47.9 \\
llama-instruct + Beam& 46.2 & 0.0 & 53.8 & 43.0 & 0.0 & 57.0 & 47.5 & 0.5 & 52.0 & 45.7 \\
llama-DPO + greedy & 77.5 & 0.0 & 22.5 & 87.0 & 0.0 & 13.0 & 83.0 & 0.4 & 16.6 & 82.6 \\
llama-DPO + temperature & 75.6 & 0.0 & 24.4 & 85.0 & 0.0 & 15.0 & 81.2 & 0.2 & 18.5 & 80.7 \\
llama-DPO + top-P & 75.0 & 0.0 & 25.0 & 82.6 & 0.0 & 17.4 & 80.0 & 0.4 & 19.6 & 79.3 \\
llama-DPO + top-K & 72.5 & 0.0 & 27.5 & 80.8 & 0.0 & 19.2 & 79.8 & 0.2 & 20.0 & 77.7 \\
llama-DPO + Beam & 74.4 & 0.0 & 25.6 & 77.4 & 0.0 & 22.6 & 77.3 & 0.2 & 22.5 & 76.4 \\
\midrule
llama-instruct + SLA & 75.0 & 1.9 & 23.1 & 82.6 & 0.2 & 17.2 & 79.3 & 2.6 & 18.1 & 79.7 \\
llama-DPO + SLA & 86.2 & 0.0 & 13.8 & 92.4 & 0.0 & 7.6 & 89.3 & 0.4 & 10.3 & \textbf{89.4} \\
\midrule
llama-DPO + SLA vs. llama-DPO & 71.9 & 0.0 & 28.1 & 77.4 & 0.0 & 22.6 & 73.0 & 1.2 & 25.8 & 74.3 \\ 
\bottomrule
\end{tabular}

\end{table*}

\subsection{Main results}
We present the performance of different approaches in~\autoref{table:main_results}. 
From the results, we can make the following observations.
First, the advanced algorithms might also select the greedy trajectory as the final output, which is the reason for the tie scenario.
Second, sampling methods such as sampling with \textit{temperature}, \textit{top-P}, and \textit{top-K} might increase the final output's diversity, but they cannot improve overall performance.
Third, \textit{DPO} is an efficient way of learning the reward signal into the model parameters. As a result, the \textit{DPO} model could outperform the baseline model under the same greedy decoding setting.
Lastly, the proposed \textit{SLA} could significantly outperform all baseline methods, and combining the \textit{DPO} technique and \textit{SLA} can further enhance the performance.
To better compare \textit{SLA}'s effect on the \textit{DPO} model, we add one more comparison of ``llama-DPO + SLA'' against ``llama-DPO + Greedy.''
The results show that although the policy model has already learned the reward preference into the parameters through \textit{DPO} training, the greedy outputs are still not optimal.
The fact that ``DPO+SLA'' could achieve an overall 74.3\% win rate against ``DPO+Greedy'' demonstrates that even for a strong and well-calibrated model, doing the looking ahead search during inference is still quite beneficial.

\subsection{Analysis on Different Search Strategy}
\begin{figure*}
    \centering
    \begin{subfigure}{0.3\textwidth} 
        \includegraphics[width=\textwidth]{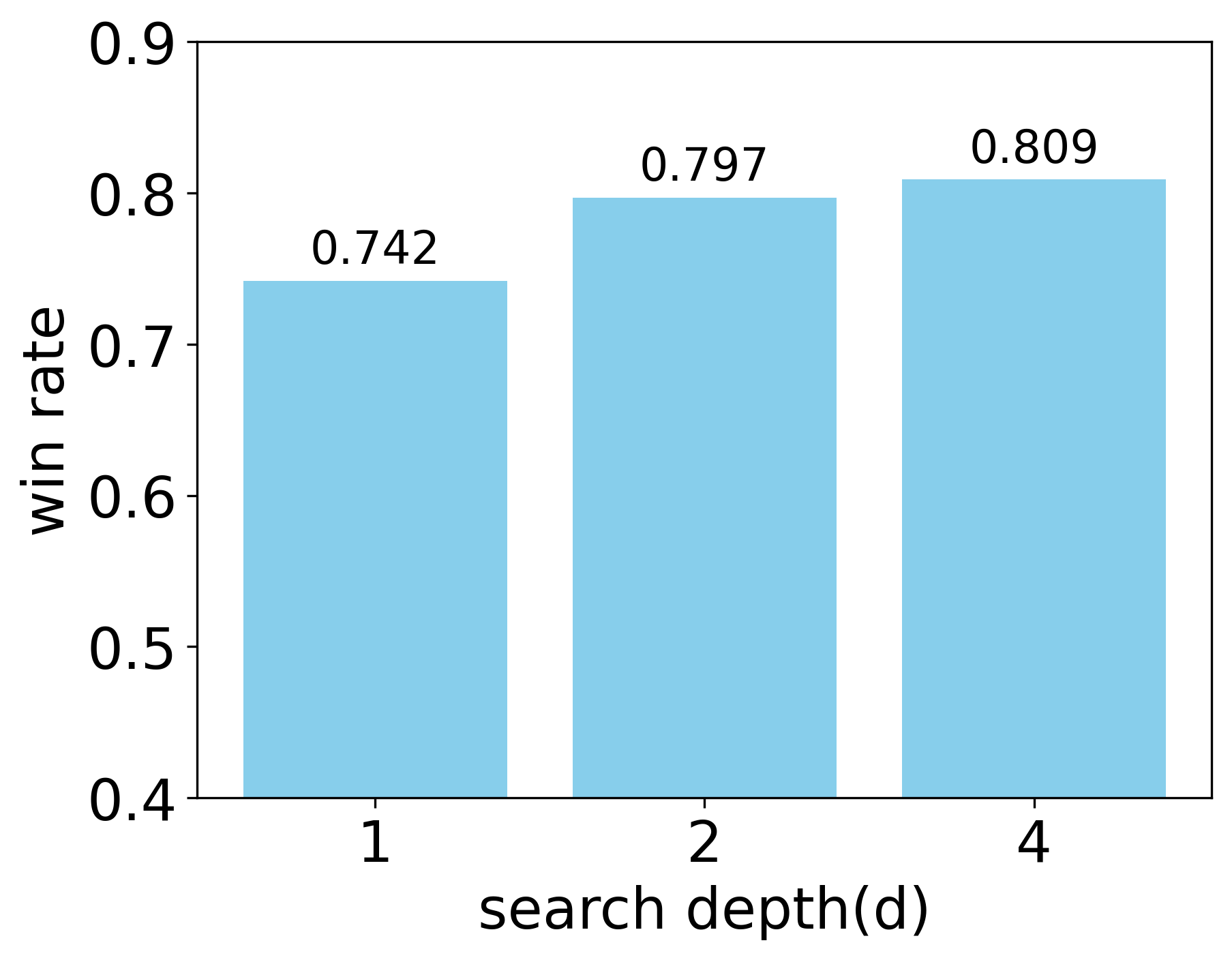} 
        \caption{Effect of the search depth.}
        \label{fig:sub1}
    \end{subfigure}
    \hfill
    \begin{subfigure}[b]{0.3\textwidth}
        \includegraphics[width=\textwidth]{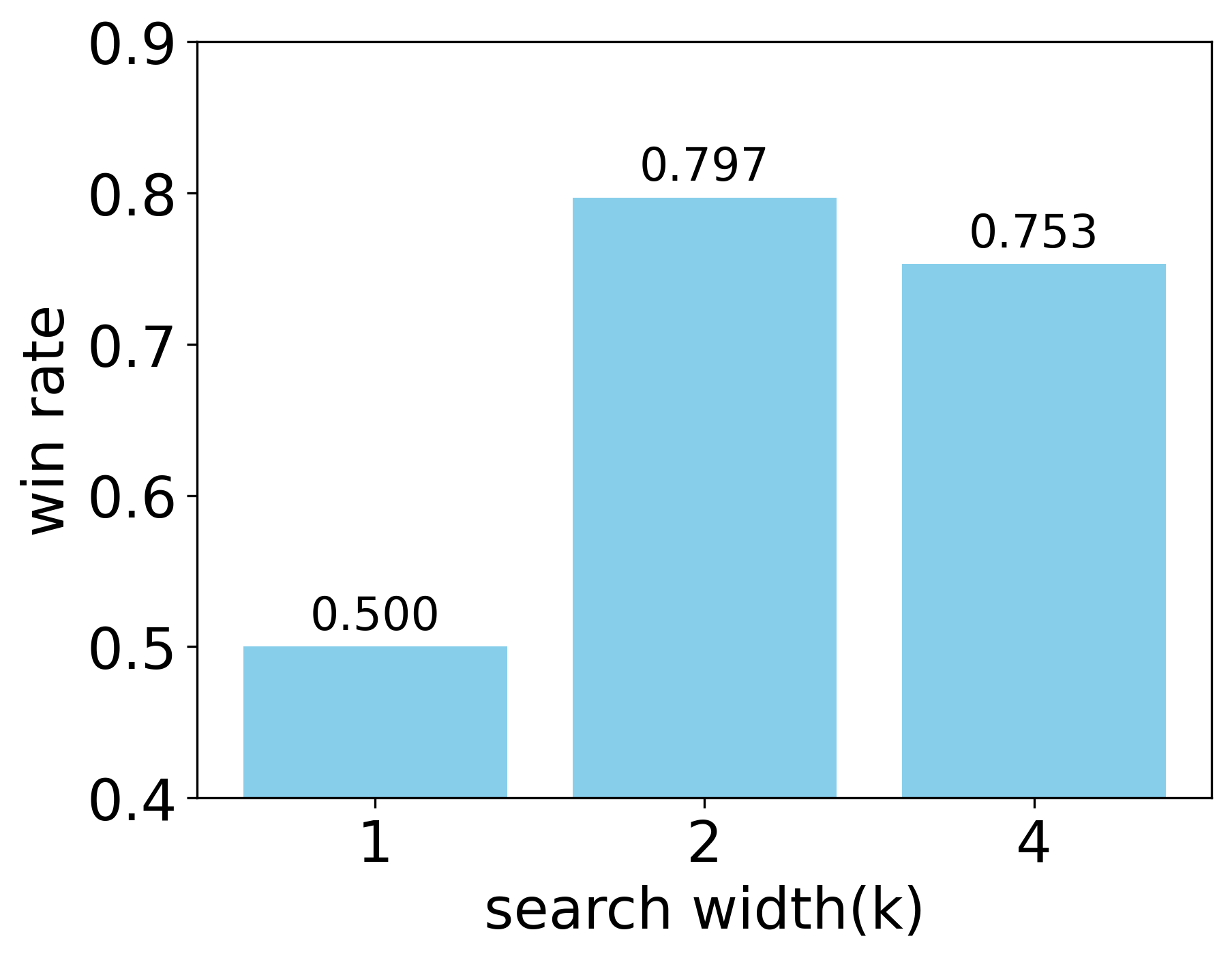} 
        \caption{effect of the search width.}
        \label{fig:sub2}
    \end{subfigure}
    \hfill
    \begin{subfigure}[b]{0.3\textwidth}
        \includegraphics[width=\textwidth]{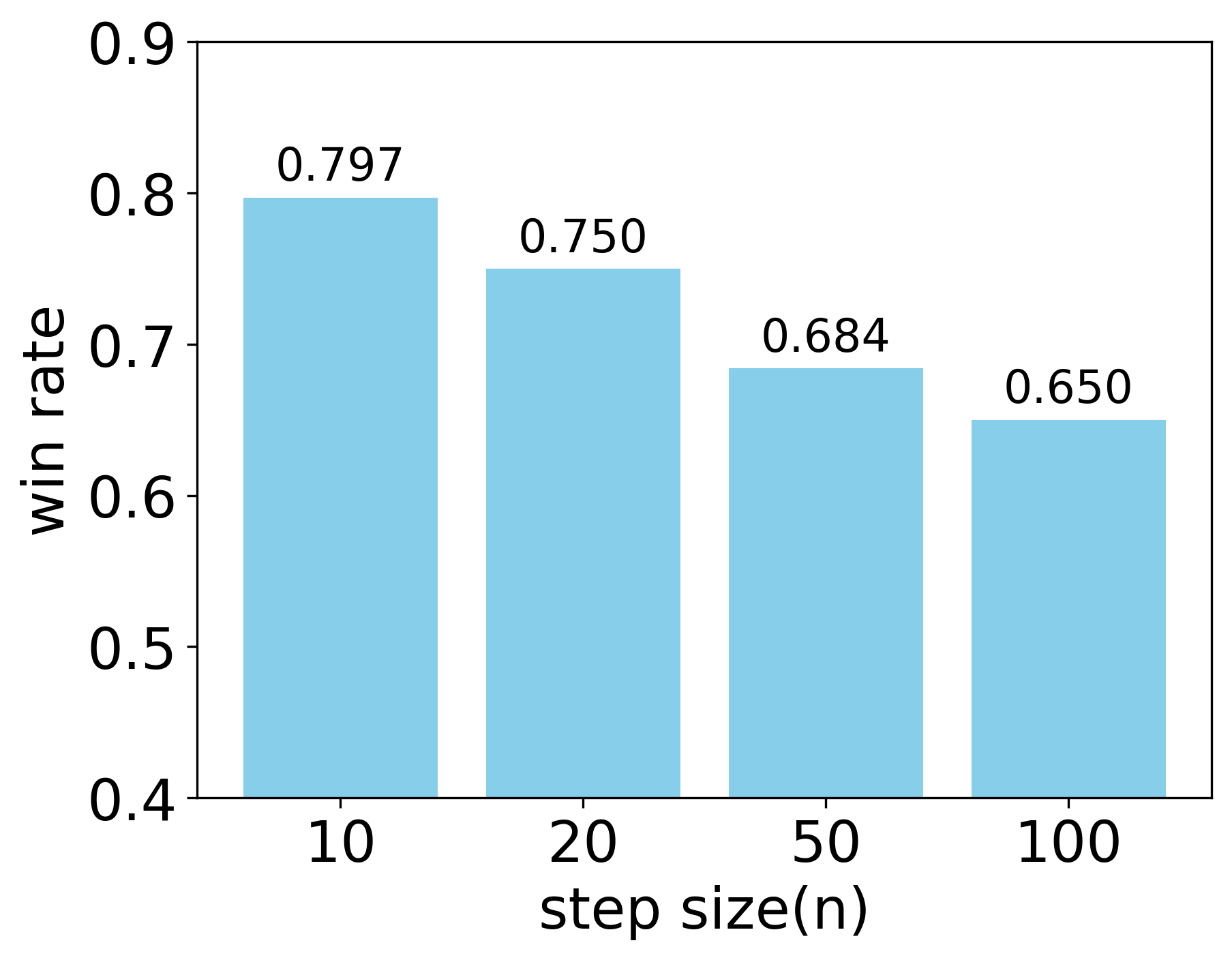} 
        \caption{effect of the search step size.}
        \label{fig:sub3}
    \end{subfigure}

    \caption{Effect of different search strategies. We set the default search depth and width to be 2 and step size to be 10. }
    \vspace{-0.15in}
    \label{fig:SLA_variations}
\end{figure*}
The streaming looking ahead algorithm has three key hyperparameters. 
The search width $k$, search depth $d$, and step size $n$.
We report the average win rate of SLA against the greedy decoding algorithm on three benchmarks and show the results in \autoref{fig:SLA_variations}.
As shown by \autoref{fig:SLA_variations} (a), the search depth could improve model performance since the model could utilize further future information and thus can make a more accurate selection.
However, results in \autoref{fig:SLA_variations} (b) show that increasing the search width might hurt the performance.
This is mainly because the TRM is not perfect, and extra noise impacts the final performance.
We leave improving the TRM performance for future research.
Lastly, in \autoref{fig:SLA_variations} (c), we show the impact of step size, from which we can see that increasing step size negatively affects final performance.
This phenomenon shows that the finer-grained search gives SLA more freedom, which further justifies the value of our framework against previous MCTS frameworks that can only use pre-defined coarse-grained steps.
Increasing the search depth linearly increases processing time and computations. 
Although the search width and step size theoretically have a minimal impact on efficiency, they still add extra computational load, consume additional GPU memory, and increase the frequency of operations like sequence copying. 
These factors can affect practical efficiency. 
Therefore, when choosing hyperparameters, one should consider the practical scenario to strike a balance between quality and efficiency.

\subsection{Analysis on Token-level Reward Modeling}
\label{sec:analysis_TRM}

\begin{figure}
    \centering
    \begin{subfigure}[b]{0.37\textwidth}
        \includegraphics[width=\textwidth]{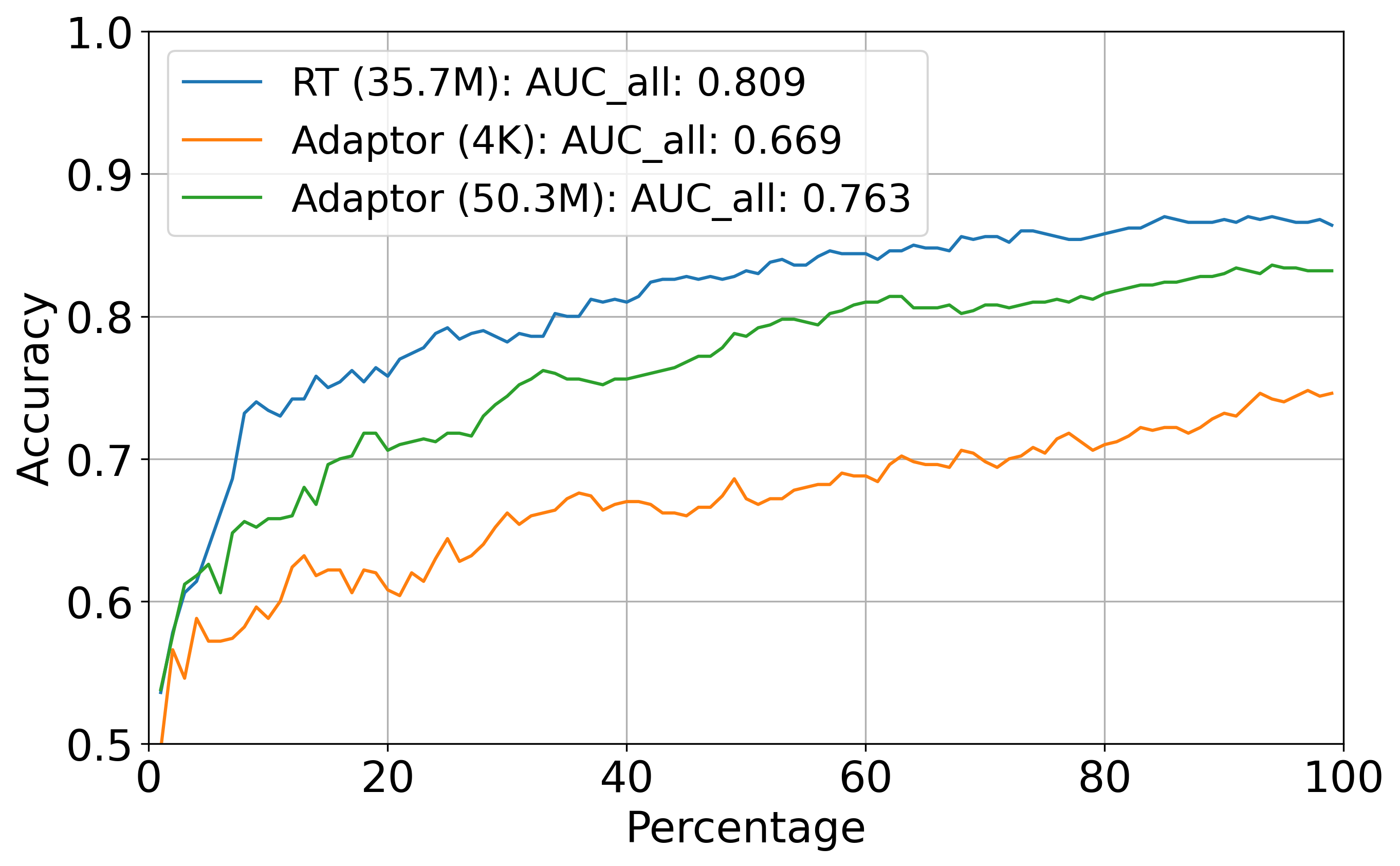} 
        \caption{Architecture design.}
        \label{fig:sub2}
    \end{subfigure}
    \begin{subfigure}{0.37\textwidth} 
        \includegraphics[width=\textwidth]{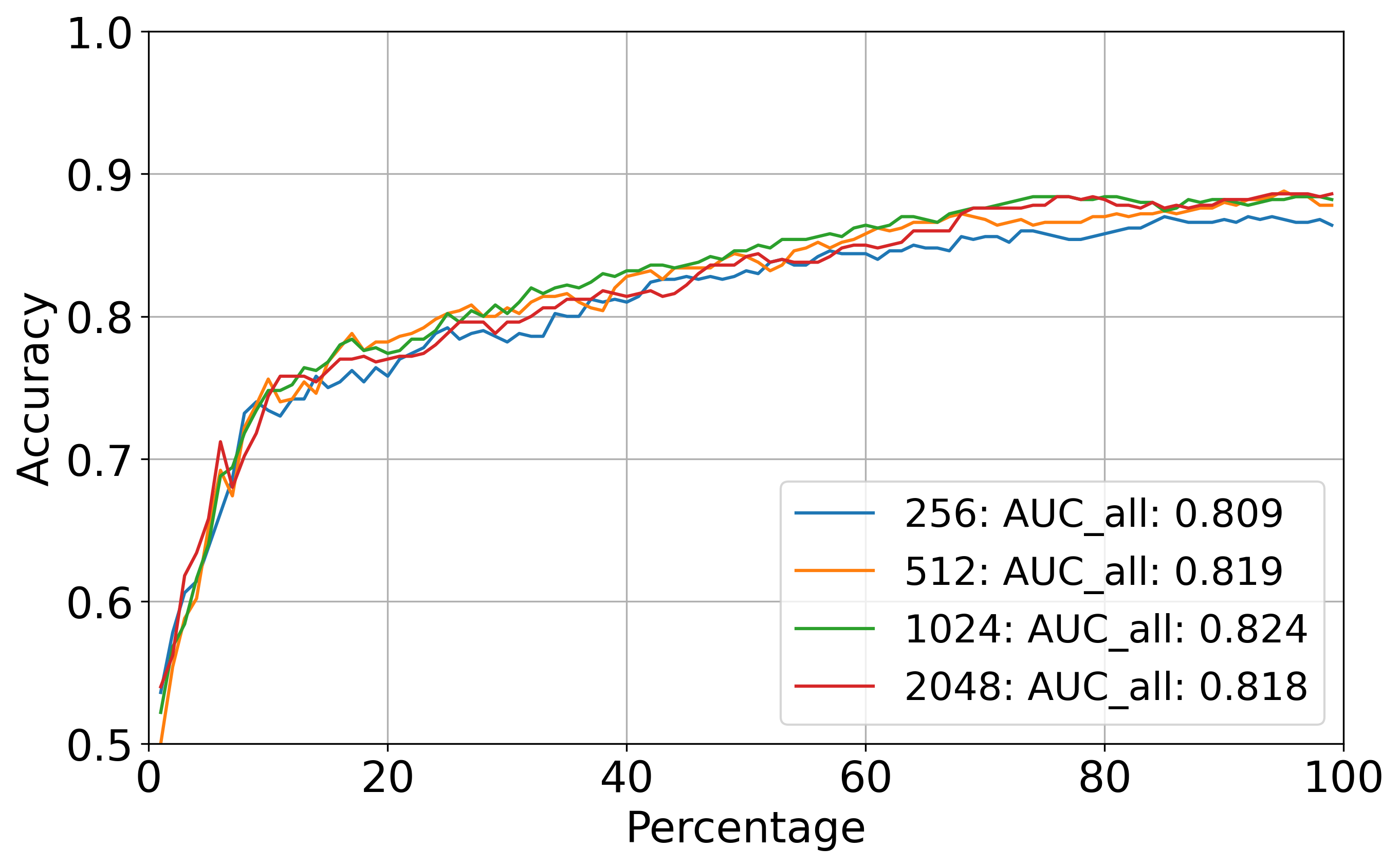} 
        \caption{Hidden dimension.}
        \label{fig:sub1}
    \end{subfigure}

    \caption{Contribution of different TRM designs. RT is the proposed architecture.  }
    \vspace{-0.1in}
    \label{fig:TRM}
\end{figure}

Another critical factor is how well the model could learn the token-level reward.
We conduct the ablation study from two perspectives: model architecture and model capacity.
Specifically, we use the widely used adapter approach as the baseline for the model architecture. 
This approach adds MLP on top of the policy model to learn the token-level reward. 
To analyze the impact of the reward channel capacity, we conducted experiments with different hidden dimensions as variations of our model.
We introduce a new evaluation metric, the Area Under Token-level Reward Curve (AuTRC), to better understand the models' TRM capability.
Given a trajectory, we can compute the agreement of the model's prediction at each token versus the ground truth.
We select the percentage as the x-axis since different trajectories might have different lengths to draw the distribution.
We can then use the area under the curve as the quantity evaluation.
We present the results in \autoref{fig:TRM}.

We compare our Reward Transformer with an additional 35.7M parameters with two variations of the adaptor approach. 
One only has a single layer and thus only has 4K additional parameters; the other one has four layers with 50.3M parameters. 
Results show that although more parameters in the adaptor help, RT could still significantly outperform the adaptor model with more parameters because the TRM requires the model to deeply understand information from all layers rather than just the abstract information in the top layer.
On the other hand, we can see that the hidden dimension does not significantly impact the TRM performance.
We could thus use a relatively small hidden representation to learn the reward as long as we evenly distribute the job into all layers.

%% file: Sections/07_conclusion.tex
\section{Conclusion}\label{sec:conclusion}


In this paper, we present Streaming Looking Ahead (\textit{SLA}), an efficient lookahead search algorithm for LLMs. 
To enable the policy to model token-level rewards, we introduce a \textit{Reward Transformer} architecture that adds a channel in each transformer module to predict the token-level reward. 
Our experiments demonstrate that SLA markedly improves streaming output quality. On three open-domain benchmarks, compared to greedy decoding, SLA achieves an overall win rate of 79.7\%, which increases to 89.4\% when combined with reinforcement fine-tuning techniques like \textit{DPO}.